\pdfoutput=1

\documentclass[11pt]{article}

\usepackage{acl}

\usepackage{times}
\usepackage{latexsym}
\usepackage{amsmath}
\usepackage{amsfonts}

\usepackage{booktabs}
\usepackage{hyperref}
\usepackage{tabularx}
\usepackage{tikz}
\usepackage{multirow}
\usepackage{xcolor}
\DeclareMathOperator*{\argmin}{arg\,min}

\usepackage[T1]{fontenc}

\usepackage[utf8]{inputenc}

\usepackage{microtype}

\usepackage{dirtytalk}

\usetikzlibrary{external}
\usepackage{todonotes}
\definecolor{dkgreen}{rgb}{0,0.6,0}
\definecolor{gray}{rgb}{0.5,0.5,0.5}
\definecolor{mauve}{rgb}{0.58,0,0.82}
\definecolor{darkblue}{rgb}{0.0,0.0,0.6}
\definecolor{cyan}{rgb}{0.0,0.6,0.6}
\definecolor{mBlue}{HTML}{4285f4}
\definecolor{mRed}{HTML}{ea4335}
\definecolor{mGreen}{HTML}{34a853}
\definecolor{mYellow}{HTML}{fbbc04}
\definecolor{mLightBlue}{HTML}{6d9eeb}
\definecolor{mLightRed}{HTML}{e06666}
\definecolor{mLightGreen}{HTML}{93c47d}
\definecolor{mLightYellow}{HTML}{ffd966}
\definecolor{archtBlue}{HTML}{9fc5e8}
\definecolor{archtTeal}{HTML}{a4d2d7}
\definecolor{archtYellow}{HTML}{ffe599}
\definecolor{archtOrange}{HTML}{f9cb9c}
\definecolor{archtCetus}{HTML}{ea9999}
\definecolor{archtOther}{HTML}{dd7e6b}
\definecolor{archtPurple}{HTML}{b4a7d6}
\definecolor{archtGray}{HTML}{eeeeee}
\definecolor{archtGreen}{HTML}{b6d7a8}
\definecolor{archtRed}{HTML}{ea9999}

\newcommand{\convdef}[1]{\textsc{Stencil}}
\newcommand{\pconvdef}[1]{\textsc{Stencil$_p$}}
\newcommand{\convdefshort}[1]{\textsc{Sten}}
\newcommand{\pconvdefshort}[1]{\textsc{Sten$_p$}}

\title{Protecting Privacy in Classifiers by Token Manipulation}

\author{
 \textbf{Re'em Harel\textsuperscript{1,2}},
 \textbf{Yair Elboher\textsuperscript{1}},
 \textbf{Yuval Pinter\textsuperscript{1}}
\\
 \textsuperscript{1}Department of Computer Science, Ben-Gurion University of the Negev, Israel \\
\textsuperscript{2}Department of Physics, Nuclear Research Center – Negev, Israel
\\
{\tt\small reemha@bgu.ac.il, yairel@bgu.ac.il, uvp@cs.bgu.ac.il}
 }

\begin{document}
\maketitle
\begin{abstract}
Using language models as a remote service entails sending private information to an untrusted provider.
In addition, potential eavesdroppers can intercept the messages, thereby exposing the information.
In this work, we explore the prospects of avoiding such data exposure at the level of text manipulation.
We focus on text classification models, examining various token mapping and contextualized manipulation functions in order to see whether classifier accuracy may be maintained while keeping the original text unrecoverable.
We find that although some token mapping functions are easy and straightforward to implement, they heavily influence performance on the downstream task, and via a sophisticated attacker can be reconstructed.
In comparison, contextualized manipulation provides an improvement in performance.
\end{abstract}

\section{Introduction}
\label{sec:intro}

Large language models (LLMs) have greatly advanced the field of NLP in recent years, exhibiting exceptional proficiency across a wide spectrum of tasks, including dependency parsing~\cite{duong-etal-2015-low}, natural language understanding~\cite{dong2019unified}, automatic question-answering~\cite{chatgpt2,chatgpt1}, machine translation~\cite{dabre2020survey}, text classification~\cite{minaee2021deep}, and many more~\cite{li-etal-2022-systematic}.
However, this success comes with potential privacy risks, as the models process vast amounts of data that might contain personal or sensitive information and may abuse or leak it.
For instance, information can be leaked by model inversion~\cite{li2017reverse}, re-identification techniques~\cite{lison-etal-2021-anonymisation,ben-cheikh-larbi-etal-2023-clinical}, exploitation of feature memorization within the LLM~\cite{carlini2021extracting}, and more. 
Offering LLMs as cloud services, such as ChatGPT~\cite{chatgpt1}, might also impose potential threats to privacy if the server exhibits a semi-honest stance, actively seeking to glean more insights from the input than is appropriate or by a possible eavesdropper intercepting the input sent to the server.

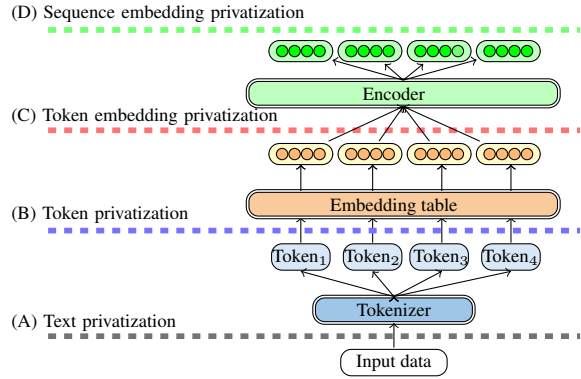
\begin{figure}[t]
    \centering
    \scriptsize
    \begin{tikzpicture}[auto, scale=0.7]
    
        \tikzstyle{embs} = [shape = rectangle, rounded corners]
        
        
        \draw [thin, rounded corners] (0.5,-4.75) rectangle (2.5,-5.25);
        \node (input) at (1.5,-5) {Input data};

        \draw[line width=0.75mm, draw=black!55, dashed] (-5,-4.5) -- (5,-4.5);
        \node[text width=9cm] at (0.75,-4.25) {(A) Text privatization};

        \draw [fill=archtBlue, thin, double, rounded corners] (0,-3.75) rectangle (3,-4.25);
         \node at (1.5,-4) {Tokenizer};

        \draw[->] (input) edge (1.5,-4.25);
        \draw [thin,fill=archtBlue!40, rounded corners] (-0.8,-2.75) rectangle (0.3,-3.25);
        \draw [thin,fill=archtBlue!40, rounded corners] (0.55,-2.75) rectangle (1.65,-3.25);
        \draw [thin,fill=archtBlue!40, rounded corners] (1.8,-2.75) rectangle (2.9,-3.25);
        \draw [thin,fill=archtBlue!40, rounded corners] (3.15,-2.75) rectangle (4.25,-3.25);
        \node (t3) at (-0.25,-3) {Token$_1$};
        \node (t4) at (1.1,-3) {Token$_2$};
        \node (t5) at (2.4,-3) {Token$_3$};
        \node (t6) at (3.7,-3) {Token$_4$};
        
        \draw[->]  (1.5,-3.75) edge (t3.south);
        \draw[->]  (1.5,-3.75) edge (t4.south);
        \draw[->]  (1.5,-3.75) edge (t5.south);
        \draw[->]  (1.5,-3.75) edge (t6.south);

        \draw[->]  (t3) edge (-0.25,-2.25);
        \draw[->]  (t4) edge (1.1,-2.25);
        \draw[->]  (t5) edge (2.4,-2.25);
        \draw[->]  (t6) edge (3.7,-2.25);
        
        \draw [fill=archtOrange, thin, double, rounded corners] (-1.25,-2.25) rectangle (4.6,-1.75) ;
        \node at (1.5, -2.05) {Embedding table};
        \draw[line width=0.75mm, draw=blue!55, dashed] (-5,-2.5) -- (5,-2.5);
        \node[text width=9cm] at (0.75,-2.2) {(B) Token privatization};

        \draw[<-]  (-0.25,-1.25) edge (-0.25,-1.75);
        \draw[<-]  (1.1,-1.25) edge (1.1 ,-1.75);
        \draw[<-]  (2.4,-1.25) edge (2.4,-1.75);
        \draw[<-]  (3.7,-1.25) edge (3.7,-1.75);
        
        \draw [thin, rounded corners, fill=yellow!25] (-0.85,-0.85) rectangle (0.35,-1.25);
              \node [matrix] (e3) at (-0.25,-1.05) {
              \draw [fill=orange!55] circle (0.75mm); & \draw [fill=orange!55] circle (0.75mm); & \draw [fill=orange!55] circle (0.75mm); & \draw [fill=orange!55] circle (0.75mm);  \\
              };
        \draw [thin, rounded corners, fill=yellow!25] (0.45,-0.85) rectangle (1.65,-1.25);
              \node [matrix] (e4) at (1.05,-1.05) {
              \draw [fill=orange!55] circle (0.75mm); & \draw [fill=orange!55] circle (0.75mm); & \draw [fill=orange!55] circle (0.75mm); & \draw [fill=orange!55] circle (0.75mm);  \\
              };
        \draw [thin, rounded corners, fill=yellow!25] (1.75,-0.85) rectangle (2.95,-1.25);
              \node [matrix] (e5) at (2.35,-1.05) {
              \draw [fill=orange!55] circle (0.75mm); & \draw [fill=orange!55] circle (0.75mm); & \draw [fill=orange!55] circle (0.75mm); & \draw [fill=orange!55] circle (0.75mm);  \\
              };
        \draw [thin, rounded corners, fill=yellow!25] (3.05,-0.85) rectangle (4.25,-1.25);
              \node [matrix] (e6) at (3.65,-1.05) {
              \draw [fill=orange!55] circle (0.75mm); & \draw [fill=orange!55] circle (0.75mm); & \draw [fill=orange!55] circle (0.75mm); & \draw [fill=orange!55] circle (0.75mm);  \\
              };
        \draw[line width=0.75mm, draw=red!55, dashed] (-5,-0.6) -- (5,-0.6);
        \node[text width=9cm] at (0.75,-0.35) {(C) Token embedding privatization};

      \draw [fill=green!25, thin, double, rounded corners] (-1.25,-0.15) rectangle (4.6,0.35) ;
        \node at (1.5, 0.1) {Encoder};

        \draw[->]  (e3) edge (1.675,-0.15);
        \draw[->]  (e4) edge (1.675,-0.15);
        \draw[->]  (e5) edge (1.675,-0.15);
        \draw[->]  (e6) edge (1.675,-0.15);
        
        \draw [thin, rounded corners, fill=green!25] (-0.85,0.7) rectangle (0.35,1.1);
              \node [matrix] (o3) at (-0.25,0.9) {
              \draw [fill=green] circle (0.75mm); & \draw [fill=green] circle (0.75mm); & \draw [fill=green] circle (0.75mm); & \draw [fill=green] circle (0.75mm);  \\
              };
        \draw [thin, rounded corners, fill=green!25] (0.45,0.7) rectangle (1.65,1.1);
              \node [matrix] (o4) at (1.05,0.9) {
              \draw [fill=green] circle (0.75mm); & \draw [fill=green] circle (0.75mm); & \draw [fill=green] circle (0.75mm); & \draw [fill=green] circle (0.75mm);  \\
              };
        \draw [thin, rounded corners, fill=green!25] (1.75,0.7) rectangle (2.95,1.1);
              \node [matrix] (o5) at (2.35,0.9) {
              \draw [fill=green] circle (0.75mm); & \draw [fill=green] circle (0.75mm); & \draw [fill=green] circle (0.75mm); & \draw [fill=green!55] circle (0.75mm);  \\
              };
        \draw [thin, rounded corners, fill=green!25] (3.05,0.7) rectangle (4.25,1.1);
              \node [matrix] (o6) at (3.65,0.9) {
              \draw [fill=green] circle (0.75mm); & \draw [fill=green] circle (0.75mm); & \draw [fill=green] circle (0.75mm); & \draw [fill=green] circle (0.75mm);  \\
              };

        \draw[->] (1.675,0.35) -> (o3);
        \draw[->] (1.675,0.35) -> (o4);
        \draw[->] (1.675,0.35) -> (o5);
        \draw[->] (1.675,0.35) -> (o6);

    \draw[line width=0.75mm, draw=green!55, dashed] (-5,1.3) -- (5,1.3);
    \node[text width=9cm] at (0.75,1.6) {(D) Sequence embedding privatization};

    \end{tikzpicture}
    
    \caption{A schematic of the various stages where differential privacy techniques can be applied in an LLM. This work focuses on level (B).}
    \label{fig:pipeline}
\end{figure}

In order to safeguard privacy, many privacy-preserving techniques have been proposed, based on the local differential privacy framework~\cite[LDP;][]{arachchige2019local}.
In this framework, the user applies a differential privacy mechanism, which can be hosted on a local server, and then sends the privatized data to the remote server.
This approach doesn't require trust from the remote server, 
and protects the data against potential eavesdroppers.
In general, any privacy mechanism can be applied at one or several components of the LLM pipeline.
\autoref{fig:pipeline} depicts these components: at the text level (\textit{text privatization}), after the tokenization process (\textit{token privatization}), after the initial embedding lookup (\textit{token embedding privatization}), or after applying several layers of the encoder (\textit{sequence embedding privatization}). 

Currently, most privacy-preserving strategies focus on incorporating noise into sequence embedding vectors. 
The rationale behind this strategy is to minimize the privacy-preserving technique's impact on the downstream task.
Specifically, most systems first obtain a sequence embedding representation, either by assuming partial access to the remote model~\cite{zhou-etal-2022-textfusion,lyu-etal-2020-differentially,qu2021natural} or by using a dedicated model to create these embeddings~\cite{li-etal-2018-towards,coavoux2018privacy,mosallanezhad-etal-2019-deep,plant-etal-2021-cape,zhou-etal-2023-textobfuscator}.
Afterwards, random noise is incorporated into the embeddings, thus concealing the original input.
However, this approach relies on partial access to the remote model, on the ability to provide input to the remote model in vector form, or on sufficient computational and memory resources on the user's end.
These are often not the case.
In addition, \citet{kugler2021invbert} showed that publishing a model's encoder along with the contextualized embeddings allows an adversary to generate data to train a decoder with a high level of reconstruction accuracy, making these approaches highly susceptible to violation of privacy.

We propose \textbf{a secure way to use LLMs without assuming access to their parameters}.
In our framework, both input and output for the privacy-providing mechanism must be given in a token sequence format, eliminating the need to intervene with the LLM's pre-training procedure or text processing. 
We focus on applying privacy preservation techniques at the token level, corresponding to layer (B) in \autoref{fig:pipeline}.

Specifically, we propose two privacy-preserving techniques based on manipulating \textbf{the input token sequence}.
The first set of techniques relies on na{\"i}ve rules of token substitution.
The second is based on leveraging contextual information to strategically replace tokens, aiming to retain as much actionable information as possible for the classifier to minimize the impact on the performance of the downstream task. 

We test these techniques both for their impact on the downstream task accuracy and for their resilience against reconstruction attacks.
We find that replacing tokens based on simple rules is easy for a knowledgeable attacker to reverse,
while manipulating tokens based on contextual information can enhance privacy without sacrificing much of the performance.\footnote{Our code is available at: \\ \url{https://github.com/MeLeLBGU/Privacy-Preserving-Token-Manipulation}.}

\section{Lossy Mapping}
\label{sec:towards}

In order to protect against potential eavesdropping by a middle party, under the assumption that the layers of LLMs are inaccessible to the local device, we start by employing several mapping functions on the tokens of the input text available at the local device.
Our initial, na{\"i}ve mapping functions introduce a random noise component that follows a specific rule: the vocabulary is partitioned into pairs of tokens $(u,v)$, or triplets $(u,v,z)$, and when encountered in an input text to be manipulated, all tokens are mapped to a single representative token of their tuple, without loss of generality $u$. 
This strategy produces outputs that are inherently ambiguous, blocking any potential eavesdroppers from recovering the original input text deterministically, given that a many-to-one mapping is not invertible.
The only available recourse for an attacker is a statistical strategy, which imposes assumptions on the properties of the input, for example that it was grammatical English text written by a speaker with high proficiency.
Indeed, even if an eavesdropper obtains full information of the privacy system, i.e.~the partition into token tuples and each tuple's representative token, each mapped sequence of length $m$ still generates a candidate set of $2^m$ or $3^m$ possible permutations (depending on tuple size) through which the attacker must search.
We will examine the practical implications of this large search space later in the section.

For our stated use case of manipulating text being input into a sequence classifier operating atop an LLM, there are two distinct scenarios depending on when we may apply our manipulation.
The first scenario involves applying the manipulation process only during the inference phase of a model trained on regular, unmanipulated text, which we will refer to as the \textsc{Test} case.
This operation mode simulates a query sent by a user to an already-trained model, such as a user interacting with ChatGPT or another model allowing only inference text interaction via user interface or an API.
In the second scenario, which we call \textsc{All}, we also apply the manipulation during the training phase, protecting sensitive information in the training data, hoping that the inference phase will now leverage the model's ability to handle manipulated input as expected and produce better results.
In this scenario the model does not inadvertently learn or memorize the sensitive data during the training process, nor does it spend learning resources on tokens never to be seen during inference, but since it is not always possible to assume its availability, we perform our experiments in both settings.

When protecting the original input data, it is essential for the mapper to have minimal impact on the performance of the downstream task, defining the fundamental trade-off in our study.
Therefore, the selection process for grouping tokens and selecting each tuple's representative token is crucial, as it aims to both minimize the mapping's effect on the downstream task and hinder the attacker's ability to uncover the original text.
We consider the following mapping functions:

\begin{table}
    \centering
    \small
    \begin{tabular}{lcccc}
        \toprule
        \textbf{Dataset} & \textbf{Mapper} & \textbf{\textsc{Test}} & \textbf{\textsc{All}} & \textbf{Unchanged}\\
        & & & & \textbf{Tokens}\\
        \midrule
        & Plain text    & 94.5\% & 94.5\% & 100\% \\
        & 2-Random      & 75.0\% & 85.0\% & 51.0\%\\
        SST2 & 3-Random & 62.0\% & 80.0\% & 34.0\% \\
        & High-freq     & 90.0\% & 91.0\% & 93.0\%\\
        & Low-freq      & 60.0\% & 78.0\% & 7.0\% \\
        \midrule
        & Plain text    & 95.0\% & 95.0\% & 100\%\\
        & 2-Random      & 75.0\% & 90.0\% & 50.0\% \\
        IMDb & 3-Random & 68.0\% & 85.0\% & 32.0\%  \\
        & High-freq     & 93.0\% & 94.0\% & 94.0\%\\
        & Low-freq      & 60.0\% & 80.0\% & 6.0\%\\
        \bottomrule
    \end{tabular}
    \caption{The mapping strategy accuracy on SST2 and IMDb datasets and the percentage of unchanged tokens after applying the mappers to the training and test sets.}
    \label{table:baseline_privacy_accuracy}
\end{table}

\paragraph{Purely random mapping} the selection of the token pairs tuples from the vocabulary and of each tuple's representative is uniformly random.
\paragraph{High-frequency mapping} token pairs are selected based on their frequency of occurrence in a tokenized corpus, such as Wikipedia~\cite{wikidump}.
This involves pairing a higher-frequency token with a lower-frequency token, with the higher-frequency token being designated as the representative.
In our mapper, given a vocabulary of even size $V$, sorted by descending frequency, each token with rank $1 \leq k \leq \frac{V}{2}$ is paired with the token of rank $k+\frac{V}{2}$.
While selecting the high-frequency token as the representative may have a lesser impact on the downstream task, it could potentially weaken the privacy-preserving characteristics, depending on the knowledge possessed by the attacker. 

\paragraph{Low-frequency mapping} the process is similar to that of the higher-frequency mapper, except that the lower-frequency token is chosen as the representative. 
Opting for less-frequent tokens as representatives can aid in preserving privacy, but it will likely harm the downstream task.

Due to the simplicity of these mapping strategies, we consider them baselines for further research and developing better, potentially language-aware strategies.
In addition, these mapping functions can easily be generalized to larger tuples, expanding the search space even further, but greatly harming downstream task performance as a result of a much more restricted active vocabulary.

\begin{figure*}[t]
    \centering
    \scriptsize
    \begin{tikzpicture}[auto,scale=1.2]
    
        \tikzstyle{embs} = [shape = rectangle, rounded corners]
        
        \node at (-0.5,4) {Input:};
        \draw [thin, rounded corners] (1,3.75) rectangle (4,4.25);
        \node (input1) at (2.5,4) {what a nice day};
        \draw [thin, rounded corners] (5,3.75) rectangle (8,4.25);
        \node (input2) at (6.5,4) {what what nice unicorn};
        \draw[->]  (input1) edge (input2);

        \node at (-0.5,1){Attacker path:};

        \draw [thin, rounded corners,fill=green!30] (0.5,1.75) rectangle (1.5,2.25);
        \node (awprob) at (1,2.4) {$p=80\%$};
         \node (w) at (1,2) {what};
         \draw [thin, rounded corners,fill=green!30] (0.5,-0.25) rectangle (1.5,0.25);
         \node (aprob) at (1,0.4) {$p=20\%$};
         \node (a) at (1,0) {a};

         \draw [thin, rounded corners,fill=red!30] (2, 2.25) rectangle (3.5, 2.75);
         \node (ww) at (2.75,2.5) {what what};
         \node (wwprob) at (2.75,2.9) {$p=80\% \times 0.1\%$};
         \draw [thin, rounded corners,fill=green!30] (2, 1.25) rectangle (3.5, 1.75);
         \node (wa) at (2.75,1.5) {what a};
         \node (waprob) at (2.75,1.9) {$p=80\% \times 99.9\%$};
         \draw[->]  (w) edge (2,2.5);
         \draw[->]  (w) edge (2,1.5);

         \draw [thin, rounded corners,fill=red!30] (2, 0.25) rectangle (3.5, 0.75);
         \node (aw) at (2.75,0.5) {a what};
         \node (awprob) at (2.75,0.9) {$p=20\% \times 10\%$};
         \draw [thin, rounded corners,fill=green!30] (2, -0.75) rectangle (3.5, -0.25);
         \node (aa) at (2.75,-0.5) {a a};
         \node (aaprob) at (2.75,-0.1) {$p=20\% \times 90\%$};
         \draw[->]  (a) edge (2, 0.5);
         \draw[->]  (a) edge (2, -0.5);

        \draw [thin, rounded corners,fill=green!30] (4.25, 2.25) rectangle (6.25, 2.75);
         \node (wab) at (5.25,2.5) {what a nice};
        \node (wwprob) at (5.25,2.9) {$p=80\% \times 99.9\% \times 90\%$};
         \draw [thin, rounded corners,fill=red!30] (4.25, 1.25) rectangle (6.25, 1.75);
         \node (wad) at (5.25,1.5) {what a is};
        \node (waprob) at (5.25,1.9) {$p=80\% \times 99.9\% \times 10\%$};
         \draw[->]  (3.3,1.6) -> (4.25,2.5);
         \draw[->]  (3.3,1.5) -> (4.25,1.5);

       \draw [thin, rounded corners,fill=green!30] (4.25, 0.25) rectangle (6.25, 0.75);
         \node (aab) at (5.25, 0.5) {a a nice};
         \node (wwprob) at (5.25,0.9) {$p=20\% \times 90\% \times 90\%$};
         \draw [thin, rounded corners,fill=red!30] (4.25, -0.75) rectangle (6.25, -0.25);
         \node (aad) at (5.25,-0.5) {a a is};
        \node (waprob) at (5.25,-0.1) {$p=20\% \times 90\% \times 10\%$};
         \draw[->]  (3.3,-0.4) -> (4.25,0.5);
         \draw[->]  (3.3,-0.5) -> (4.25,-0.5);
        
        \draw [thin, rounded corners] (7, 2.25) rectangle (10, 2.75);
         \node (wab) at (8.5,2.5) {what a nice unicorn};
         \node (wwprob) at (8.5,2.9) {$p=80\% \times 99.9\% \times 90\% \times 1\%$};
         \node (wwprob) at (11,2.5) {$P=7.1928\%$};
         \draw [thin, rounded corners,fill=green!100] (7, 1.25) rectangle (10, 1.75);
         \node (wad) at (8.5,1.5) {what a nice day};
         \node (wwprob) at (11,1.5) {\textbf{$P=64.7352\%$}};
        \node (waprob) at (8.5,1.9) {$p=80\% \times 99.9\% \times 90\% \times99\%$};
         \draw[->]  (6.3,2.5) -> (7,2.5);
         \draw[->]  (6.3,2.3) -> (7,1.5);

       \draw [thin, rounded corners] (7, 0.25) rectangle (10, 0.75);
         \node (aab) at (8.5, 0.5) {a a nice day};
         \node (wwprob) at (8.5,0.9) {$p=20\% \times 90\% \times 90\% \times 50\%$};
         \node (wwprob) at (11,0.5) {$P=8.1\%$};

         \draw [thin, rounded corners] (7, -0.75) rectangle (10, -0.25);
         \node (aad) at (8.5,-0.5) {a a is unicorn};
         \node (waprob) at (8.5,-0.1) {$p=20\% \times 90\% \times 90\% \times 50\%$};
         \node (wwprob) at (11,-0.5) {$P=8.1\%$};
         \draw[->]  (6.3,0.5) -> (7,0.5);
         \draw[->]  (6.3,0.3) -> (7,-0.5);

    \end{tikzpicture}
    
    \caption{Schematic overview of the proposed heuristic oracle attacking scenario path over trying to reconstruct the sentence ``what a nice day'' which is remapped to ``what what nice unicorn''. 
    The red boxes indicate that the probability (presented above the box) of the candidate is low enough to be dropped in the next step, while the green boxes are the candidates that will be expanded in the next step.}
    \label{fig:attacker}
\end{figure*}
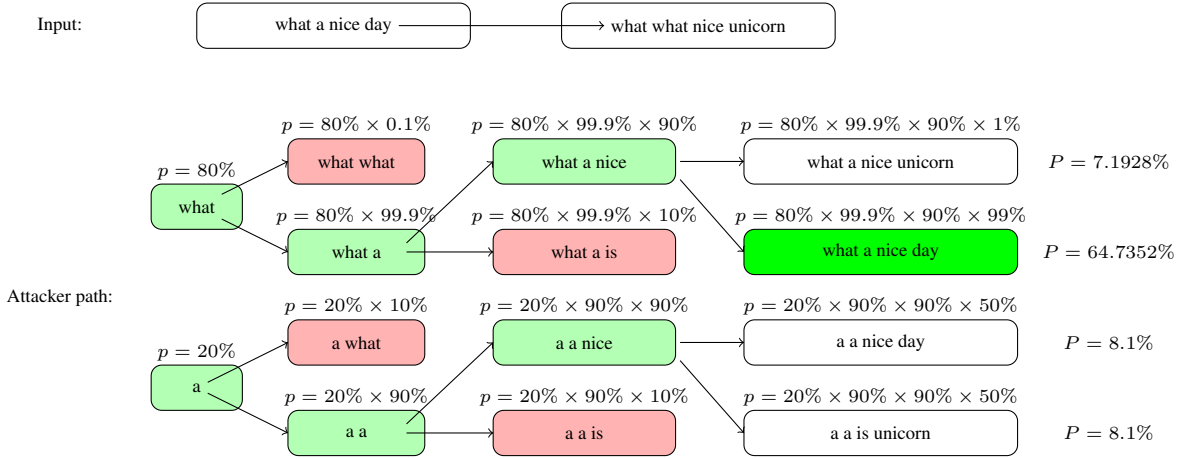

\subsection{Task Performance}
To assess the impact of the baseline models on downstream task performance, we use two datasets for sequence classification: SST2~\cite{socher-etal-2013-recursive} and IMDb~\cite{maas-etal-2011-learning}. 
The base model chosen was RoBERTa~\cite{liu2019roberta}, a state-of-the-art encoder language model known for its strong performance in sequence classification tasks.
In \autoref{table:baseline_privacy_accuracy}, we present the results of four baselines on the two datasets, compared with the null mapping results labeled \say{Plain text}.
Perhaps unsurprisingly, the high-frequency baseline achieved the highest accuracy, most likely due to the fact that retaining high-frequency tokens while removing low-frequency ones results in a relatively small number of tokens altered in the datasets.
In both datasets this number is roughly 6\%, compared with low-frequency mapping's complement of 94\% and with the randomly-selected sets' 50\% and 67\%, giving a correlative relationship between this number and the performance level:
the fewer tokens are altered, the better the model performs.
This effect is much more pronounced when only the test set is affected, and the model is dealing not only with loss of information but also with out-of-distribution behavior.
In absolute terms, we find it remarkable that this alteration of a non-negligible portion of tokens causes only a 1--2 percentage point reduction in performance for the IMDb dataset and still under 5 points for SST2.

\begin{table}
\centering
\small
\begin{tabular}{llll}
\toprule
\textbf{Mapper} & & \textbf{Text} &  \\
\midrule
 Plain Text                  & \textbf{no} & \textbf{apparent} & \textbf{joy}\\
 2-Random                  &  his & buffers & University\\
 High-freq                 &  no & apparent & joy\\
 Noise(150)                &  non & evident & joyful \\
 \convdefshort{}(9, 0.8)   &  No & evident & joyful\\
 \pconvdefshort{}(9, 1.0)  &  apparent & No & joyful \\

\bottomrule
\end{tabular}
\caption{Examples of the privatized textual sequences obtained with different
privacy-preserving techniques.}
\label{table:detailed}
\end{table}

In \autoref{table:detailed}, we present an example of the outcome of applying the 2-Random and the High-freq privatization techniques on a random phrase (\say{no apparent joy}) from the SST2 dataset.
As expected, the 2-random baseline produces a random sequence of words, whereas the high-frequency mapper leaves the phrase unchanged as the tokens in the original sequence are frequent.

\subsection{Brute-force Attacker}
\label{ssec:brute}

Although the many-to-one mapping function introduces some form of protection against data leakage, in practice, reconstructing the original text might be relatively straightforward under certain circumstances.
In particular, if an \say{oracle} attacker has access to the token pairings, it can theoretically determine the original text from the pool of $2^m$ possible permutations by applying a generative LLM such as GPT~\cite{gpt2} and picking the most probable sequence.
However, generating and evaluating all $2^m$ permutations is impractical even for small values of $m$ due to the computational complexity involved. 
To mitigate this challenge, alternative approaches, such as employing heuristics or utilizing statistical methods, can be explored to narrow down the potential candidates for the original text.

To cope with this task, we describe a heuristic approach to reducing the search space based on \textbf{beam search}~\cite[][\S 11.3.1]{eisenstein2018natural} and \textbf{nucleus sampling}~\cite{holtzman2019curious}.
In each step of the process, candidates are generated based on the prefixes of tokens that were produced in the previous steps.
In the case of token pairs, each prefix sequence is followed by one of two candidate tokens for the next step based on the known (oracle) token pair that the observed representative token belongs to.
Unlike conventional beam search, where a fixed number of candidates is retained following each step, we opt for a dynamic approach inspired by nucleus sampling, made possible since the scores for each of the two tokens reflect a generative probabilistic process where the relative probability of each interim token sequence on the beam can be estimated and used for dropping highly unlikely sequence prefixes.
This means that the number of candidates remaining on the beam varies at each step, adapting to their likelihood and ensuring flexibility in the selection process.
We estimate the likelihood of each candidate prefix using a language model.\footnote{\url{https://github.com/simonepri/lm-scorer}}
After all prefixes on the beam have been scored, we remove the least probable candidates such that the total probability of the remaining candidates exceeds a certain threshold $\pi$ set by computational constraints but maintaining discoverability.
Since the probability of a sequence cannot exceed that of its prefix, the process guarantees that complete sequences that are likely are not being discarded before getting the chance to be fully generated.
Overall, this process effectively eliminates highly unlikely candidates, dramatically reducing the search space during its application and streamlining the computational efforts.

This process is illustrated in \autoref{fig:attacker}.
The \say{oracle} attacker gains access to the remapped words: \texttt{(what,a)}$\rightarrow$ \texttt{a}, \texttt{(nice, is)} $\rightarrow$ \texttt{nice}, \texttt{(day, unicorn)} $\rightarrow$ \texttt{unicorn}. 
In the first step, two initial candidates (\texttt{what} and \texttt{a}) are generated based on the first observed token (\texttt{what}).
Following the described process, each prefix is evaluated via an LLM to determine its probability, for instance, the probability of \texttt{what} being the first word is 80\% when considering the possible set \{\texttt{[s] what}, \texttt{[s] a}\}.
This process is repeated, and the candidates with low probability are removed, such that the total probability of the remaining candidates is above 85\%, as indicated by the red boxes. 
Finally, the probability of the sequence \texttt{what a beautiful day} is the highest, thus the \say{oracle} attacker returns it as the inferred original text.
We note that the low-frequency and high-frequency mappers, despite their differences in representative token selection, will demonstrate equivalent safeguarding mechanisms against this attacker since the attacker does not factor in the choice of the representative token and examines all potential candidates in its effort to uncover the original text.

\begin{table}
\centering
\small
\begin{tabular}{lcccc}
\toprule
  \textbf{Dataset} & \textbf{Mapper}  & \textbf{MRR} & \textbf{Pr@5} & \textbf{Edit dist}\\
   & & ($\downarrow$) & ($\downarrow$) & ($\uparrow$)
    \\ \midrule
        & 2-Random  & 0.89 & 0.97 & 1.32  \\
SST2    & 3-Random  & 0.81 & 0.92 & 1.35 \\
        & High-freq & 0.86 & 0.98 & 1.33 \\
\midrule
        & 2-Random  & 0.48  & 0.59  & 1.60  \\
IMDb    & 3-Random  & 0.45  & 0.53  & 1.70 \\
        & High-freq & 0.63  & 0.72   & 1.60 \\
\bottomrule
\end{tabular}
\caption{The three random mappings' capability of preserving privacy against an \say{oracle} attacker. Edit distance is calculated at the token level.}
\label{table:baseline_privacy_attacker}
\end{table}

\subsection{Resilience Against Reconstruction Attacks}
In \autoref{table:baseline_privacy_attacker}, we present the outcomes of the attacker's endeavors to reveal the original text from the three techniques: 2-Random, 3-Random, and High-freq (equivalent to Low-freq for a knowledgeable attacker).
We report the mean reciprocal rank (MRR) of the correct sequences, the rate of the actual input sequence ranking among the top 5 predictions (Pr@5), and the token-level edit distance between the produced top prediction and the original sequence.
The relative success of the mappers in thwarting the oracle attacks on the IMDb dataset compared to SST2 can be attributed to the average token sequence length ($\bar{m}$), which is 65 and 12, respectively.
As sequence length increases, the attacker's task of uncovering the original text becomes more challenging.

Our results indicate that the na{\"i}ve baselines are overly simplistic and allow an easy and straightforward reconstruction, even within a vast search space (although attacker knowledge of the mapping specifications is required).
In cases where performance on the task remains close to that of unmapped text, the recovery price is too high to neglect.
Having said that, the computational complexity of applying the na{\"i}ve baselines is relatively low, and the greatly reduced active vocabulary brings great savings in parameter budgets, which embedding tables often dominate. 
In a less powerful attack environment, this would make them an efficient choice for preserving privacy on low-resource devices.
We expect future work on more principled many-to-one static mappings would be able to improve both task performance and resilience to attackers, while work on attack strategies can present challenges hitherto unseen.

\section{\convdef{} Privacy Preservation}
\label{sec:fancy}
In the context of protecting privacy within NLP practices, a widely adopted approach for implementing local differential privacy involves introducing a controlled level of \emph{noise} into different components of the model, effectively concealing the original input. 
These components may include sequence embeddings, token embeddings, or the tokens themselves~\cite{mosallanezhad-etal-2019-deep,feyisetan2020privacy,lyu-etal-2020-differentially,qu2021natural,zhou-etal-2022-textfusion}.
However, in essence, the success of models in most NLP tasks is primarily attributed to their effective utilization of contextual information.
Moreover, our study focuses on token-level privacy preservation, i.e.,~we assume that the parameters of the LLMs are inaccessible, 
making the importance of contextual information more pronounced.
Therefore, a fundamental limitation associated with incorporating noise is the exclusion of contextual information when defining the noise.
This omission may hinder the potential benefits contextual details can offer for maintaining the performance of the downstream tasks.

Given this limitation, we propose a new privacy preservation technique, which we call \convdef{}.\footnote{This term hails from numerical analysis~\cite{spotz1995high}, where it denotes a computation that involves the surrounding values.}
With this technique, a mapped token in a sequence \say{absorbs} information from adjacent tokens to form a new context-aware token, effectively concealing the original token while retaining information beneficial for maintaining task performance.

In order to generate the new contextualized token $t_k \rightarrow t'_k$, we first retrieve an embedding vector representation of the neighborhood, of size $n+1$, containing the tokens $t_{i}, \forall i\in\{k-n/2\dots k+n/2\}$ using some embedding lookup table $\mathbf{E}\in\mathbb{R}^{V\times d}$, which can be trained independently in a preliminary step or obtained from an available model such as the target model itself.
We then subject the $n+1$ embedding vector representations to a weighted transformation and incorporate them to form a new \say{quasi-embedding} vector $\sum_{i=k-n/2}^{k+n/2} f_i\cdot\mathbf{E}[t_i]$.
Finally, we return the token $t'_k$ that is closest to the quasi-embedding vector in the embedding space, based on cosine-similarity or euclidean distance computation, as an output. 
To further enhance privacy, we ensure that the new token is different from the original one.
Formally, the process can be defined as follows:
\begin{equation}\label{eq:convdef}
     t'_k = \argmin_{t_j \in \mathcal{V}} \left\| \mathbf{E}[t_j] - \sum_{i=k-\frac{n}{2}}^{k+\frac{n}{2}} f_i\cdot\mathbf{E}[t_i] \right\|,
\end{equation}
where $\mathcal{V}$ is the vocabulary and $f_i$ is the weighted transformation function of the tokens such that $\sum_{i=k-\frac{n}{2}}^{k+\frac{n}{2}}f_i=1$. 

The level of privacy enhancement and its impact on the downstream task by employing the \convdef{} method can be managed by adjusting the window size and the properties of the weighted function $f$. 
In our study, we use the gaussian smoothing function as the weighted function.
Consequently, the standard deviation, $\sigma$, plays a crucial role in the performance and amount of privacy achieved.

As a baseline for our proposed technique, we adopt \citet{qu2021natural}'s proposed privacy-preserving technique.
In contrast to our proposed technique, this approach does not consider context but rather incorporates random noise into token embeddings to enhance privacy.
The random noise is obtained by multiplying a sample from a Gamma distribution $\Gamma (d, 1/\eta)$ and a uniform sample from a unit hypersphere, where $\eta$ corresponds to the amount of noise introduced to the original token and $d$ is the dimension of the embedding space.

We note that the most time-intensive operation in both \convdef{} and noise-based techniques involves searching for the closest token to the perturbed quasi-embedding vector, while all other operations are negligible in comparison.
Overall, the average computational cost per token is 0.005 seconds on two 16-core 3.2 GHz AMD EPYC 7343 Milan processors.

\subsection{Downstream Task Performance}
\begin{table}
\setlength{\tabcolsep}{4pt}
\small
\centering
\begin{tabular}{lcrrr}
\toprule
\textbf{Dataset} & \textbf{Mapper} & \textbf{\textsc{Test}} & \textbf{\textsc{All}} & \textbf{Pr@5} \\
 & & ($\uparrow$) & ($\uparrow$) & ($\downarrow$) \\
\midrule
\multirow{5}{1em}{SST2} & Plain Text                    & 94.5\% & 94.5\%  & - \\

& Noise(100)                    & 80.0\% & 87.8\%  & 70.0\% \\
& Noise(150)                    & 83.0\% & \textbf{90.0\%}  & 75.0\% \\
& \convdefshort{}(9, 0.8)  & 83.5\% & 89.3\%  & 49.0\% \\
&  \pconvdefshort{}(9, 1.0)        & \textbf{85.0\%} & 87.0\%  & \textbf{0.0\%} \\
\midrule
\multirow{5}{1em}{IMDb} & Plain Text                    & 95.0\% & 95.0\% & -\\
& Noise(100)                    & 89.0\% & 92.6\% & 86.0\% \\
& Noise(150)                    & 90.0\% & \textbf{93.5\%} & 90.0\% \\
 & \convdefshort{}(9, 0.8)  & \textbf{90.2\%} & 93.1\% & 67.0\% \\
 & \pconvdefshort{}(9, 1.0)        & 89.7\% & 92.4\% & \textbf{0.0\%} \\

 \midrule
\multirow{5}{1em}{QNLI} & Plain Text  & 88.1\% & 88.1\%             & -\\
& Noise(100)                          & 80.0\% & 84.0\%          & 93.0\% \\
& Noise(150)                          & \textbf{81.1\%} & \textbf{84.4\%} & 93.0\% \\
& \convdefshort{}(9, 0.8)             & 74.8\% & 83.1\% & 54.0\% \\
& \pconvdefshort{}(9, 1.0)            & 67.9\%          & 82.5\% & \textbf{0.0\%} \\
\bottomrule
\end{tabular}
\caption{The best results achieved by the \convdef{} mapper and the noise mapper considering the Test and All cases on the SST2, IMDb, and QNLI datasets. Pr@5 represents the average token hit managed by the nearest-neighbor attacker.}
\label{table:stencil_acc_Best}
\end{table}

To evaluate the impact of the \convdef{} method and of the noise-based technique on model performance, we repeat the methodology outlined in \S\ref{sec:towards}: we use RoBERTa as the base model and for the word embedding lookup table; SST2 and IMDb as the datasets; and the two distinct application cases: manipulating tokens on inference data only (\textsc{Test}), and applying the technique during the training phase as well (\textsc{All}).
However, as these privacy techniques exhibit a realistic case, we also test it on an encoder-decoder model T5-small~\cite{raffel2020exploring} on the QNLI task from the GLUE dataset~\cite{wang2019glue}. 
As in \newcite{raffel2020exploring}, we concatenate the question and its corresponding sentence to form a single sequence that serves as the input, while the target prediction is either \say{entailment} or \say{not\_entailment}, thus forming a classification task.

We report two distinct manipulations based on \convdef{}.
The first approach follows the process described in (\ref{eq:convdef}), where the weighting function $f_i$ is derived from a gaussian smoothing with a standard deviation of $\sigma=0.8$ and the number of adjacent tokens considered is set to nine (four from each side, as well as the target token).
To preserve model performance, the tokenizer and embedding lookup table used to derive the new tokens were sourced directly from the model being trained.
In the second approach, which we call punctuated \convdef{}, denoted \pconvdef{}, we exclude the target token from the computation of the quasi-embedding vector in (\ref{eq:convdef}) by setting $f_k$ to zero.
This exclusion significantly diminishes the attacker's ability to reconstruct the original token at the expense of performance.
The standard deviation we consider for this approach is $\sigma=1.0$, with a window width of nine.
For the baseline approach, we report the two best $\eta$ values: $\eta=100, 150$.

The results are presented in \autoref{table:stencil_acc_Best}.
The best accuracy is obtained with Noise ($\eta=150$) in the \textsc{All} case, where higher values of $\eta$ yield smaller noise.
This comes at great cost in discoverability, to be presented in \S\ref{ssec:nnr}.

Compared to the sentiment analysis tasks (SST2 and IMDb), the QNLI task presents greater challenges, primarily due to the complex logical connections required for the model to discern entailment between the given sentence and question.
Therefore, despite its instance sizes being very similar to those of IMDb (62 vs. 65), the fact that noise-based perturbations disrupt contextual and semantic information leads to a significant decrease in the model's ability to discern the logical connections between the parts of the input.
This results in a more pronounced performance degradation compared to the long-sequenced IMDb on the \textsc{Test} case.
In contrast, training the model on the noisy data (the \textsc{All} setup) proves effective in overcoming this effect, leading to improved results for T5-small.

\begin{table}[ht!]
\setlength{\tabcolsep}{4pt}
\centering
\small

\begin{tabular}{lcrrr}

\toprule
\textbf{Dataset} & \textbf{Mapper} & \textbf{\textsc{Test}} & \textbf{\textsc{All}} & \textbf{Pr@5} \\
 & & ($\uparrow$) & ($\uparrow$) & ($\downarrow$) \\
\midrule

\multirow{10}{1em}{SST2}& Plain text & 94.5\% & 94.5\% & --- \\
& \convdefshort{}(9, 0.2)           & 87.0\% & 91.9\% & 75.6\% \\
& \convdefshort{}(9, 0.6)           & 85.0\% & 91.0\% & 75.1\% \\
& \convdefshort{}(9, 0.8)           & 83.0\% & 89.2\% & 49.5\% \\
& \convdefshort{}(9, 1.0)           & 83.2\% & 86.4\% & 18.4\% \\

& \pconvdefshort{}(9, 0.2)          & 65.0\% & 70.0\% & 0.0\% \\
& \pconvdefshort{}(9, 0.6)          & 83.0\% & 85.0\% & 0.0\% \\
& \pconvdefshort{}(9, 0.8)          & 85.0\% & 86.0\% & 0.0\% \\
& \pconvdefshort{}(9, 1.0)          & 86.0\% & 87.0\% & 0.0\% \\

\midrule
\multirow{10}{1em}{IMDb}& Plain text & 95.0\%    & 95.0\% & --- \\
& \convdefshort{}(9, 0.2)           & 91.6\%    & 93.9\% & 94.0\% \\
& \convdefshort{}(9, 0.6)           & 89.3\%	& 93.5\% & 91.0\% \\
& \convdefshort{}(9, 0.8)           & 90.1\%	& 93.1\% & 67.0\% \\
& \convdefshort{}(9, 1.0)           & 86.5\%	& 91.4\% & 32.0\% \\

& \pconvdefshort{}(9, 0.2)          & 70.0\% & 77.0\% & 0.0\% \\
& \pconvdefshort{}(9, 0.6)          & 89.6\% & 91.4\% & 0.0\% \\
& \pconvdefshort{}(9, 0.8)          & 89.2\% & 92.0\% & 0.0\% \\
& \pconvdefshort{}(9, 1.0)          & 89.7\% & 92.4\% & 0.0\% \\

\midrule
\multirow{10}{1em}{QNLI}& Plain text & 88.1\%    & 88.1\% & --- \\
& \convdefshort{}(9, 0.2)           & 81.6\%    & 84.7\% & 93.0\% \\
& \convdefshort{}(9, 0.6)           & 81.3\%	& 83.5\% & 88.2\% \\
& \convdefshort{}(9, 0.8)           & 74.8\%	& 83.1\% & 54.1\% \\
& \convdefshort{}(9, 1.0)           & 69.7\%	& 81.4\% & 35.3\% \\

& \pconvdefshort{}(9, 0.2)          & 53.2\% & 72.0\% & 0.0\% \\
& \pconvdefshort{}(9, 0.6)          & 63.4\% & 82.0\% & 0.0\% \\
& \pconvdefshort{}(9, 0.8)          & 64.5\% & 82.2\% & 0.0\% \\
& \pconvdefshort{}(9, 1.0)          & 67.9\% & 82.5\% & 0.0\% \\
\bottomrule
\end{tabular}
\caption{The \convdef{} mappings accuracy with different values of $\sigma$ with a window size of 9, considering the \textsc{Test} and \textsc{All} cases on the SST2, IMDb and QNLI datasets. Pr@5 represents the average token hit managed by the nearest-neighbor attacker.}
\label{table:stencil_acc_sigma}
\end{table}

\begin{table}[ht!]
\setlength{\tabcolsep}{4pt}
\centering
\small
\begin{tabular}{lcrrr}
\toprule
\textbf{Dataset} & \textbf{Mapper} & \textbf{\textsc{Test}} & \textbf{\textsc{All}} & \textbf{Pr@5} \\
 & & ($\uparrow$) & ($\uparrow$) & ($\downarrow$) \\
\midrule
\multirow{12}{1em}{SST2} & Plain text              & 94.5\%  & 94.5\% & --- \\
                        & \convdefshort{}(5, 0.2) & 85.2\%	& 91.5\% & 75.0\%\\
                        & \convdefshort{}(7, 0.2) & 85.4\%  & 91.2\% & 75.0\%\\
                        & \convdefshort{}(9, 0.2) & 87.2\%	& 91.9\% & 75.0\%\\
                        & \convdefshort{}(11, 0.2)& 86.0\%  & 91.2\% & 75.0\%\\

                        & \pconvdefshort{}(5, 0.2) & 79.0\%	& 82.0\% & 0.0\%\\
                        & \pconvdefshort{}(7, 0.2) & 73.0\%  & 75.0\% & 0.0\%\\
                        & \pconvdefshort{}(9, 0.2) & 65.2\%	& 70.0\% & 0.0\%\\
                        & \pconvdefshort{}(11, 0.2)& 67.0\%  & 67.0\% & 0.0\%\\
\midrule
\multirow{12}{1em}{IMDb} & Plain text              & 95.0\% & 95.0\% & ---\\
                        & \convdefshort{}(5, 0.2) & 91.2\% & 93.9\% & 94.0\% \\
                        & \convdefshort{}(7, 0.2) & 91.4\% & 93.9\% & 94.0\% \\
                        & \convdefshort{}(9, 0.2) & 91.6\% & 93.9\% & 94.0\% \\
                        & \convdefshort{}(11, 0.2)& 91.8\% & 93.9\% & 94.0\%\\

                        & \pconvdefshort{}(5, 0.2) & 84.5\%	& 88.9\% & 0.0\%\\
                        & \pconvdefshort{}(7, 0.2) & 77.3\%  & 83.7\% & 0.0\%\\
                        & \pconvdefshort{}(9, 0.2) & 70.2\%	& 77.0\% & 0.0\%\\
                        & \pconvdefshort{}(11, 0.2)& 73.0\%  & 75.0\% & 0.0\%\\

\midrule
\multirow{12}{1em}{QNLI} & Plain text              & 88.1\% & 88.1\% & ---\\
                        & \convdefshort{}(5, 0.2) & 81.7\% & 82.3\% & 93.0\% \\
                        & \convdefshort{}(7, 0.2) & 82.0\% & 85.4\% & 93.0\% \\
                        & \convdefshort{}(9, 0.2) & 81.6\% & 85.1\% & 93.0\% \\
                        & \convdefshort{}(11, 0.2)& 81.4\% & 85.1\% & 93.0\%\\

                        & \pconvdefshort{}(5, 0.2) & 60.4\%	& 78.9\% & 0.0\%\\
                        & \pconvdefshort{}(7, 0.2) & 56.2\% & 75.7\% & 0.0\%\\
                        & \pconvdefshort{}(9, 0.2) & 53.2\%	& 72.0\% & 0.0\%\\
                        & \pconvdefshort{}(11, 0.2)& 52.8\% & 69.2\% & 0.0\%\\
                        
\bottomrule
\end{tabular}
\caption{The \convdef{} mappings accuracy with different values of the window size with $\sigma=0.2$, considering the \textsc{Test} and \textsc{All} cases on the SST2, IMDb and QNLI datasets. Pr@5 represents the average token hit managed by the nearest-neighbor attacker.}
\label{table:stencil_acc_window}
\end{table}

In \autoref{table:detailed}, we present an example of the outcome of applying \convdef{}, \pconvdef{}, and the noise mapper on a random phrase from the SST2 dataset.
The noise mapper with a value of $\eta=150$ introduces negligible noise, thus producing a similar sequence to the original one.
The \convdef{}-based techniques also produce a similar sequence, although \pconvdef{} swaps the positions of some tokens as a direct result of excluding the target token from the obfuscation process.

\subsection{Nearest-neighbor Reconstruction}
\label{ssec:nnr}

An attacker can potentially exploit the fact that these techniques utilize contextualized tokens and the selection of the nearest token as the quasi-embedding vector~\cite{qu2021natural}.
Specifically, given the new, perturbed token $t'$, the attacker can obtain the embedding vector representation $\mathbf{E}[t']$.
Afterward, the attacker can calculate the cosine similarity between $\mathbf{E}[t']$ and the other embedding vector representations ($\mathbf{E}[t]$ where $t \in \mathcal{V} \setminus \{t'\})$ and statistically determine the original token.
Hence, to test the resilience of these techniques against token inversion attacks, we implement the described attacker and report whether the original token was found to be one of the nearest five (Pr@5).

The success rate of the attacker for the four techniques is presented in \autoref{table:stencil_acc_Best}.
While the minor alterations in the original tokens contributed to performance improvement in the noise mapper, it is found to be highly vulnerable to simple reconstruction attacks. 
Taking into account both accuracy and resilience against reconstruction attacks, the \convdef{} method demonstrates better results, with a marginal trade-off in performance.

\subsection{Impact of Window Size and $\sigma$}
\label{ssec:ablation}

To better understand the impact of the window size and the value of $\sigma$ on the accuracy and resilience against reconstruction attack, we conduct tuning experiments for these values.
In \autoref{table:stencil_acc_sigma}, we present the accuracy results of the \convdef{} method applied to the SST2, IMDb, and QNLI datasets, with varying values of $\sigma$ while keeping the window size constant at 9.
Low values of $\sigma$ imply prioritizing the central token. 
Hence, the new token will likely be similar to the original token, yielding the highest accuracy results but rendering it more susceptible to reconstruction attacks.
In contrast, opting for a higher value of $\sigma$ will reduce the accuracy results while providing better resilience against the nearest-neighbor reconstruction attacks.

In \autoref{table:stencil_acc_window}, we present the accuracy results of \convdef{} on the datasets, examining the impact of different window sizes while maintaining a constant value of $\sigma=0.2$. 
Given that the average number of tokens in the SST2 dataset is below 10, incorporating 11 neighbors is likely not advantageous, making a window size of 9 yield optimal results.
Similarly, for the IMDb and QNLI datasets, optimal results are achieved when considering 11 neighbors. 
Nevertheless, in comparison to the variable values of $\sigma$, the window size exerts a lesser influence on the accuracy of the downstream task and demonstrates no impact on privacy. 
This limited effect of the window size, in contrast to the influence of $\sigma$, stems from the primary influence of the original token on the downstream task. 
Consequently, considering more neighbors has a diminished impact.

\section{Conclusion}
\label{sec:conclusion}

In this paper, we propose several token manipulation methods to preserve privacy under the assumption that the model parameters are inaccessible.
We first introduce four mappers that offer advantages compared to existing privacy-preserving techniques. 
These mappers operate independently of the LLM and the specific downstream task, resulting in a high degree of versatility. 
Additionally, their computational complexity is relatively low, making them efficient choices for privacy preservation on local, low-resource devices. 
However, these mappers harm the performance of the downstream tasks and can be easily reconstructed by a knowledgeable attacker. 

The second mapper class we propose is based on utilizing contextualized information to maintain performance while obfuscating the original input text.
This technique achieves higher privacy measures and has less impact on the downstream task, which makes it more applicable for cases where the downstream task is important.
Nevertheless, opting for different weighted functions, such as ones based on a trained model, can further help improve both accuracy and privacy.

An inherent problem with existing privacy-preserving techniques is their inability to maintain linguistic properties such as grammar and readability (as seen in \autoref{table:detailed}) that are crucial for the performance of the model.
Therefore, an additional avenue we plan to explore is application of these and similar rules in differential privacy techniques. 
For instance, following the application of random perturbations to an embedding vector, instead of simply returning the nearest token to the perturbed vector, one could consider returning a token with similar syntactic attributes, such as part of speech, or verbs with similar causative meanings or stable subcategorization frames.

Lastly, our experiments were limited to classification tasks in the English language. 
In future research, we intend to explore the effectiveness of these methods in generative tasks, across languages, and in multilingual settings.

\section*{Limitations}

We demonstrated the privacy achieved by our methods empirically under one attacking scenario.
Further comprehensive testing or mathematical proofs would enhance our understanding of the extent of privacy achieved.

An additional limitation of our proposed mechanism is the unchanged sentence length.
This imposes a privacy breach in which an author who prefers writing longer or shorter sentences can be re-identified even when introducing random perturbations.
Hence, another avenue in this research is reducing the amount of tokens by introducing, for example, a stride parameter to the \convdef{} family of mappers.
This parameter will determine how often tokens will be output, thus reducing the amount of tokens.

\section*{Acknowledgments}
We thank Niv Gilboa and Michael Elhadad for discussions on the fundamentals of this work.
We thank the anonymous reviewers for their comments.
This research was funded by the Israeli Ministry of Science and Technology (Grant 22/5451). 
Re'em Harel was supported by the Lynn and William Frankel Center for Computer Science.

\bibliography{anthology,nlpprivacy}
\bibliographystyle{acl_natbib}

\end{document}